\PassOptionsToPackage{numbers,compress}{natbib}
\documentclass{article}
\usepackage{wrapfig}
\usepackage[main, final]{neurips_2026}
\usepackage[utf8]{inputenc}
\usepackage[T1]{fontenc}
\usepackage{hyperref}
\usepackage{url}
\usepackage{booktabs}
\usepackage{amsmath,amssymb,amsfonts,mathtools}
\usepackage{nicefrac}
\usepackage{microtype}
\usepackage{xcolor}
\usepackage{graphicx}
\usepackage{subcaption}
\usepackage{multirow}
\usepackage{enumitem}
\usepackage{xspace}
\usepackage{algorithm}
\usepackage{algpseudocode}
\usepackage{tikz}
\usetikzlibrary{arrows.meta,positioning,fit,calc,shapes.geometric}
\usepackage{siunitx}

\setlist[itemize]{leftmargin=*,topsep=2pt,itemsep=2pt}
\setlist[enumerate]{leftmargin=*,topsep=2pt,itemsep=2pt}

\newcommand{\method}{\textsc{Scout}\xspace}
\newcommand{\ghost}{\textsc{Ghost}\xspace}

\DeclareMathOperator*{\argmax}{argmax}

\title{\method: Planning under Occlusion via Object-Centric World Model Rollouts}

\author{%
  Amir Hossain Raj \qquad Xuesu Xiao \\
  George Mason University \\
  \texttt{\{araj20, xiao\}@gmu.edu}
}

\begin{document}

\maketitle

\begin{abstract}
Robots manipulating in cluttered spaces must often plan from partial views where foreground objects occlude the target, surrounding geometry, and usable free space. Reactive policies often miss these hidden constraints, while exhaustive search becomes computationally prohibitive in dense clutter. We present \textsc{SCOUT}, an occlusion-aware planning framework that bridges high-level deliberation with grounded physical rollouts. \textsc{SCOUT} leverages an object-centric scene representation to reconstruct hidden geometry and derive a relational planning state capturing blocker hierarchies and accessibility corridors. To navigate the vast action space in clutter, an LLM proposes structured action candidates which are then rigorously evaluated via Monte Carlo Tree Search (MCTS) using a learned action-conditioned world model. Evaluations on `Reveal' and `Placement' tasks show that \textsc{SCOUT} outperforms Vision-Language-Action (VLA) policies, embodied agents, and 3D world models, with significant gains in long-horizon tasks requiring reasoning over occluded geometry. We further validate the framework’s robustness through real-world robotic manipulation in a cluttered shelf environment.
\end{abstract}

\section{Introduction}

Manipulation in cluttered, occluded spaces is fundamentally difficult because the objects and free space that matter most are often not directly visible \citep{dogar2012nonprehensile,wong2013activeoccluded}. A target may be buried behind front-row blockers in a shelf, reachable only through a narrow corridor that is itself occluded or only after several intermediate rearrangements \citep{danielczuk2019mechanical,bejjani2021occlusion}. Reasoning about these unobservable spaces to manipulate in clutter requires a robot to move beyond purely reactive visual feedback, necessitating deliberation across possible environment and object layouts and their intricate relationships among each other.

\begin{figure}[t]
    \centering
    \includegraphics[width=\textwidth]{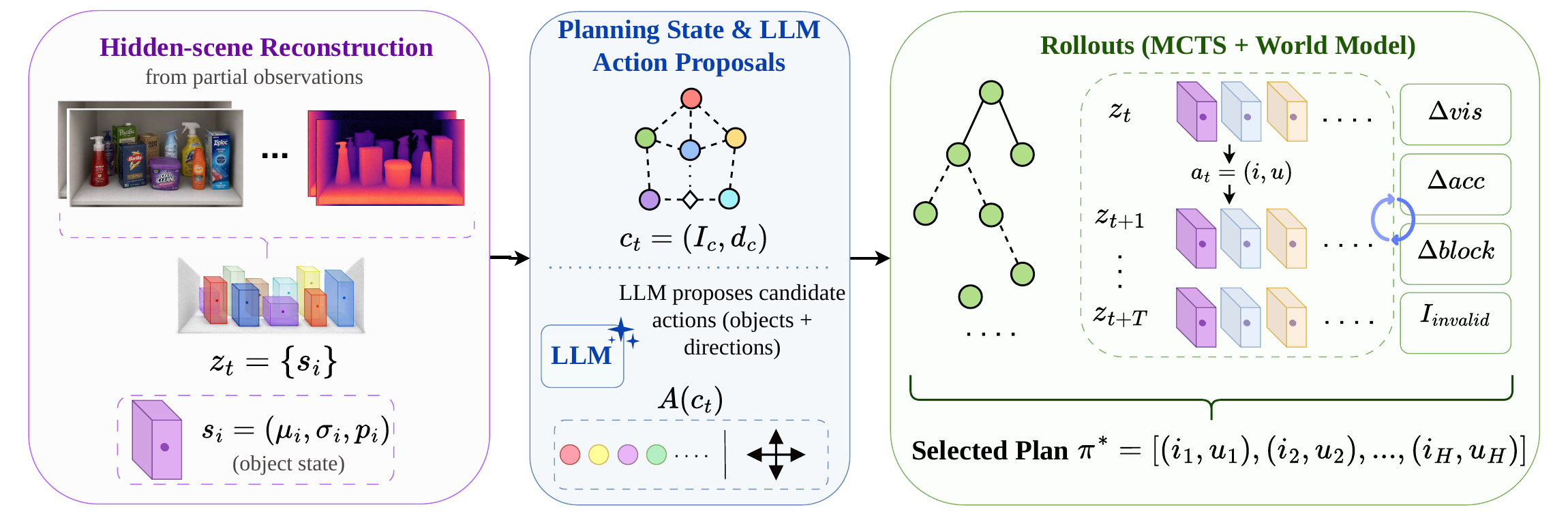}
    \caption{Overview of SCOUT. From partial RGB-D observations, GHOST reconstructs an object-centric  hidden scene representation, which is converted into a relational planning state. An LLM generates structured action proposals, and MCTS with a learned world model evaluates simulated rollouts to guide selection of a multi-step plan.}
    \label{fig:overview}
\end{figure}

A natural way to enable such deliberation is to plan in belief space, where the robot represents unobserved scene structure and reasons over possible hidden configurations \citep{kaelbling2013beliefspace}. For occluded manipulation, such a belief must capture more than the mere existence of hidden objects. The planner must also reason about which object is likely to block the target and where free space may exist around it. This suggests an explicit scene representation that captures object-level geometry and spatial structure, rather than only symbolic hidden-state hypotheses. Moreover, representing the hidden scene is only part of the problem. To manipulate in clutter, the robot must also reason about how that scene representation will evolve under candidate actions. It therefore needs a world model that predicts how an action reveals the target, creates free space, or simply changes the occlusion pattern \citep{hafner2019planet}. 

Recent progress in vision-language and foundation-model-based robotics has improved action selection from observed images and led to strong results in semantic grounding, skill selection, program synthesis, and visually guided planning \citep{ichter2023saycan, liang2023codeaspolicies, singh2023progprompt, huang2023voxposer, nasiriany2024pivot, liu2024okrobot}. These methods are effective when the relevant objects and affordances are visible, showing that large models can provide useful high-level structure for robotic decision making. World models and search offer a complementary route by supporting imagination-based planning over candidate futures \citep{schrittwieser2020muzero, hafner2023dreamerv3, wu2022daydreamer, yang2024unisim, bar2025nwm}, and recent language-guided search methods combine Large Language Models (LLMs) with search to improve long-horizon decision making \citep{chang2024lgmcts, dainese2024gifmcts}. However, these approaches still largely operate over the currently observed scene or over abstract state representations that do not explicitly capture the hidden 3D structure created by physical clutter, which are essential for occluded manipulation.  

We therefore present \method, a planning framework for manipulation under occlusion that combines hidden-scene reconstruction, action-conditioned prediction, and LLM-based deliberation. Given a small number of RGB-D observations from the frontal viewpoints, our reconstruction module, GHOST (Geometry-aware Hallucination for Occluded Scene Topology), produces an object-centric scene representation capturing geometry and occupancy beyond the observed view. A learned world model predicts how this representation evolves under discrete object-level actions \citep{hafner2019planet}. An LLM proposes structured action candidates, and a Monte Carlo Tree Search (MCTS)~\cite{browne2012survey} module evaluates simulated rollouts under the learned world model. The next primitive is selected via grounded search, and planning repeats from the updated state.
We evaluate the method on cluttered manipulation tasks in which targets are initially occluded, partially observable, and require multi-step rearrangement to access \citep{dogar2012nonprehensile, wong2013activeoccluded}. In these settings, we find that planning over reconstructed scene structure and simulated futures is more effective than relying on visible-only reasoning or one-shot language predictions alone.

\textbf{Our contributions are:}
\begin{itemize}\parskip0.25em 
    \item A structured hidden-scene representation for occluded manipulation from sparse frontal RGB-D views.
    \item A planning method that reasons over this representation using a learned world model and LLM-guided search.
    \item Simulation and real-robot validation on cluttered multi-step rearrangement tasks.
\end{itemize}

\section{Related Work}

\paragraph{Planning under Occlusion with Hidden-Scene Models.}
Planning under occlusion is classically a belief-space problem \citep{kaelbling1998planning}. In embodied AI, recent work has addressed hidden targets and incomplete maps through semantic priors, diffusion-based planning, and uncertainty-aware task-and-motion planning. PONI predicts potential locations of unseen targets from partial semantic maps \citep{ramakrishnan2022poni}, while value-guided diffusion planning explicitly targets navigation under partial observability through multi-step candidate generation and best-plan selection \citep{zhang2024valueguided}. TAMPURA and related partially observable TAMP methods reason under action uncertainty using symbolic belief representations and information-gathering strategies \citep{curtis2024tampura, kaelbling1998planning}, with extensions incorporating LLM-guided state estimation over hidden objects \citep{kim2026cocotamp}. Occluded planning further depends on reconstructing hidden 3D structure from limited observations. Prior work in semantic scene completion and open-world 3D scene understanding reconstructs geometry and semantics beyond the field of view \citep{cao2022monoscene, li2023voxformer, nguyen2024open3dis, agarwal2024scenecomplete}. Spatially aware robot foundation models show that explicit geometric alignment improves manipulation quality and generalization \citep{qu2025spatialvla, li2025spatialforcing}, while recent work leverages robot trajectories and test-time adaptation to improve action semantics and value estimation \citep{chen2025robo2vlm, ziakas2026vita}. In contrast, we address planning under occlusion in cluttered 3D scenes by reasoning over a reconstructed hidden scene and using an action-conditioned world model to evaluate how different actions reveal previously unobserved structure.

\paragraph{Language and vision foundation models for robotics.}
Foundation-model approaches have enabled robots to select skills, synthesize code, and plan trajectories from language and images \citep{ichter2023saycan, liang2023codeaspolicies, singh2023progprompt, huang2023voxposer}. PIVOT shows that iterative visual prompting can substantially improve spatial decision-making by repeatedly refining proposals rather than committing to one-shot predictions \citep{nasiriany2024pivot}. OK-Robot demonstrates how open-knowledge components can be composed into a robust mobile-manipulation pipeline, while recent analysis highlights the importance of spatial grounding and operational detail when integrating such modules \citep{liu2024okrobot}. Our method differs from these works in two key respects: it explicitly reasons over a hidden-scene belief rather than only visible observations, and it uses the language model as a \emph{search prior}, while the world model and MCTS perform physical evaluation and final action selection.

\paragraph{World models and search for embodied planning.}
Learned world models have proved effective for planning and policy improvement in games and control, including MuZero, Dreamer, and DayDreamer \citep{schrittwieser2020muzero, hafner2023dreamerv3, wu2022daydreamer}. More recent simulators and generative world models expand this perspective to real-world video and navigation \citep{yang2024unisim, bar2025nwm}. Language-guided search has also been explored for object rearrangement and code world-model generation \citep{chang2024lgmcts, dainese2024gifmcts}. Our goal is not to simply attach an LLM to search. Instead, the LLM narrows the object-action space using the relational planning state, while MCTS and the learned world model evaluate candidate futures and select the final plan from grounded rollout evidence.
\section{Method}

We address long-horizon manipulation under partial observability by combining an object-centric scene representation, a learned world model, and an LLM that generates structured action proposals for planning. Our formulation explicitly separates (i) action proposal generation from the relational planning state, from (ii) physical feasibility and multi-step outcome prediction, handled by the learned world model. Fig.~\ref{fig:overview} provides an overview of \method, from hidden-scene reconstruction and LLM-guided action proposals to world-model rollouts and final plan selection. The complete execution flow of the system is detailed in Algorithm~\ref{alg:scout}.

\subsection{Problem Formulation}
\label{sec:problem_formulatio}
We consider manipulation under partial observability, where an agent must plan a sequence of object-level actions from a limited set of observations. Each instance is defined by a hidden scene state $s$, a goal specification $g$, and a finite action budget $T$. The agent receives a set of RGB-D observations $o = \{o^{(1)}, \dots, o^{(V)}\}$ captured from frontal viewpoints, providing partial visibility of objects, free space, and access corridors.
The objective is to compute a finite-horizon action sequence $\mathbf{a}_{1:T} = (a_1, \dots, a_T)$, where $a_t \in \mathcal{A}$, that achieves the task goal. 
The action space $\mathcal{A}$ consists of object-centric manipulation primitives $a = (i, u)$, where $i$ indexes the manipulated object and $u \in \{\texttt{shift\_left}, \texttt{shift\_right}, \texttt{shift\_back}, \texttt{pull\_forward}\}$.
In the initial-view protocol, the scene is reconstructed once
from $o$, and subsequent planning states are updated using
the world model without re-observation.

We consider four task families instantiated from a shared distribution of cluttered shelf scenes: \textbf{Reveal}, \textbf{Reveal\_Multi}, \textbf{Place}, and \textbf{Place\_Multi}. Each task defines a binary success condition evaluated at the end of the sequence. Reveal tasks require the target to be visible and accessible, while placement tasks require the object to lie within a goal region, with Reveal\_Multi and Place\_Multi being the counterparts over multiple objects. 

The system operates over a reconstructed scene representation $z_0$ obtained from observations, and a derived planning state $x_0$ capturing task-relevant structure (Sec.~3.2). A learned world model predicts state transitions $z_{t+1} = f_{\phi}(z_t, a_t)$ (Sec.~3.3), enabling simulation of action-conditioned future states. The planner can be viewed as a mapping $\Pi: (o, g) \rightarrow \mathbf{a}_{1:T}$, which uses the reconstructed scene representation, LLM action proposals, and world-model predictions to evaluate candidate action sequences up to horizon $T$. The objective is to select a plan that maximizes task success:
\begin{equation}
\pi^* = \argmax_{\mathbf{a}_{1:T} \in \mathcal{A}^T}
\; \mathbb{E}[\mathrm{Success}(z_T, g)].
\end{equation}

\subsection{Object-Centric Scene Representation}

\begin{figure}[t]
    \centering
    \includegraphics[trim={0 3.9cm 0 4cm}, width=\textwidth]{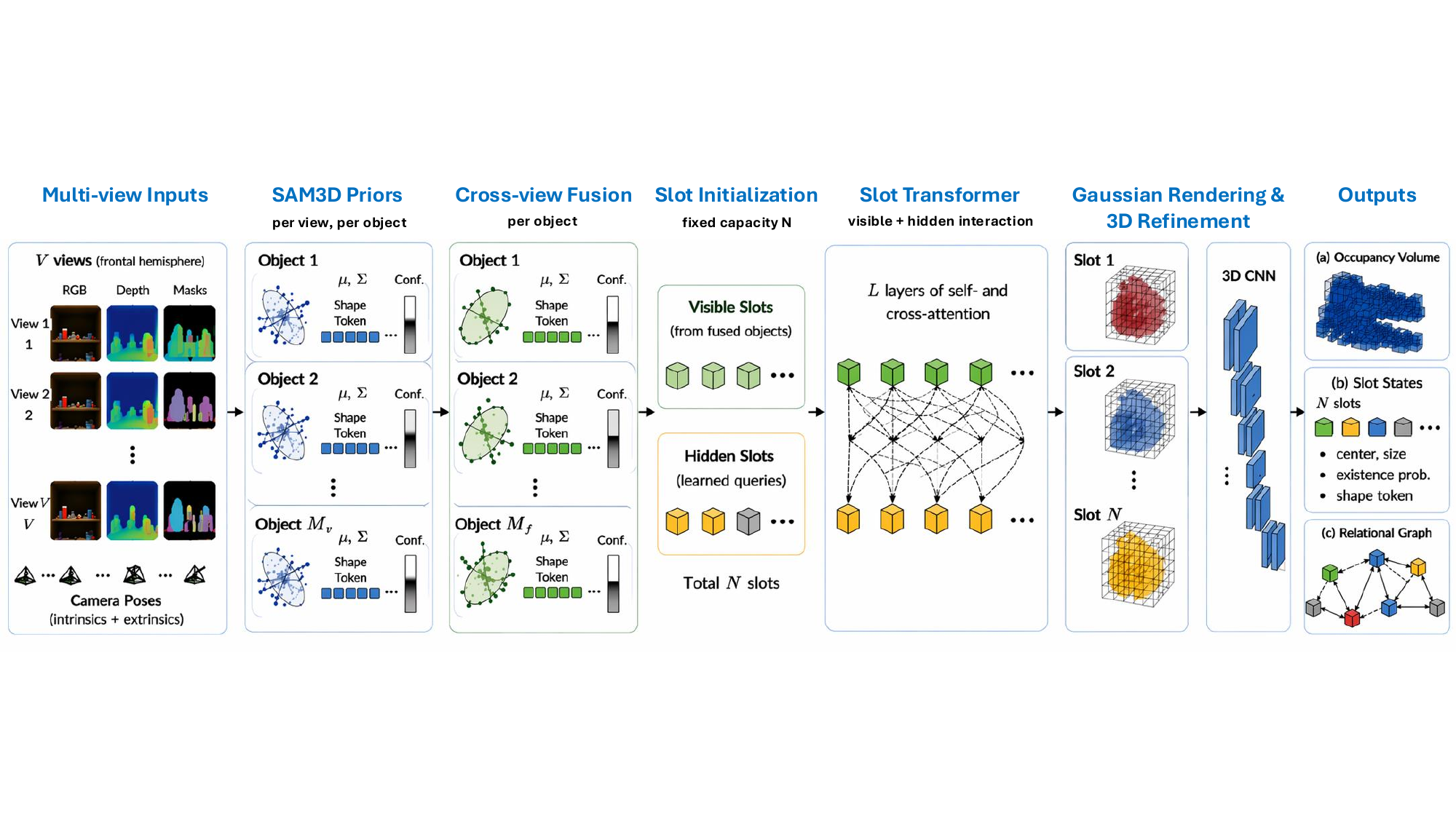}
    \caption{GHOST scene representation pipeline. Multi-view RGB-D observations are fused into object-centric slots, refined via slot interactions, and rendered into a unified 3D scene representation.}
    \label{fig:GHOST}
\end{figure}

\setlength{\intextsep}{4pt}
\setlength{\columnsep}{10pt}
\begin{wrapfigure}{R}{0.28\textwidth} 
  \vspace{-18.5pt}
  \centering
  \includegraphics[trim={0.75cm, 0.75cm, 0.75cm, 0.75cm},width=0.28\textwidth]{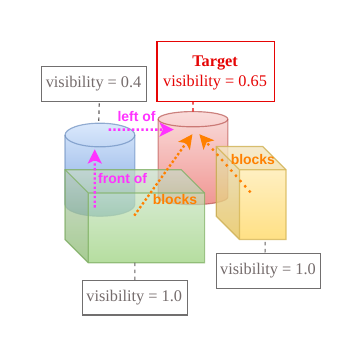}
  \caption{Relational planning state derived from object-centric slots.}
  \vspace{+0.2pt}
  \label{fig:relational_graph}
\end{wrapfigure}
Given observations $o$, we construct an object-centric state representation $z_t = \ghost(o)$, where $z_t = \{ s_i^t \}_{i=1}^{K}$ and each slot $s_i^t = (\mu_i^t, \sigma_i^t, p_i^t)$ encodes a hypothesized object instance with center $\mu_i^t \in \mathbb{R}^3$, 3D size $\sigma_i^t \in \mathbb{R}^3$, and existence probability $p_i^t \in [0,1]$.

As shown in Fig.~\ref{fig:GHOST}, GHOST takes RGB-D views from the front-facing viewpoints, together with camera intrinsics and extrinsics. For each view, SAM3D ~\cite{sam3d2025} provides object-level priors from segmented object regions, including coarse 3D geometry estimated from the mask and depth, a confidence score indicating how reliable the object hypothesis is, and a compact shape token that summarizes the object's local appearance and geometry in a learned feature vector. These per-view object hypotheses are fused across views into object-level estimates with uncertainty, producing visible slots for objects supported by observations. To account for occluded structure, GHOST also maintains a fixed-capacity set of hidden slots initialized as learned queries. The visible and hidden slots are jointly refined using a slot transformer with self-attention and cross-attention, allowing observed objects and latent hypotheses to exchange spatial context.

The refined slots are rendered through a Gaussian-based 3D refinement module to produce an occupancy volume, updated slot states, and a relational graph. To support reasoning about task structure, we derive a relational planning state $x_t = \mathcal{P}(z_t, g)$ from the reconstructed slot set $z_t$ and the goal $g$, as illustrated in Fig.~\ref{fig:relational_graph}. Each of the $K$ slots contains 7 values: a 3D center, a 3D size, and an existence probability. The resulting state $x_t$ includes (i) the target object index, (ii) blocker relationships indicating which objects obstruct access to the target, (iii) visibility indicators, and (iv) coarse spatial relations. These relations are computed deterministically from slot geometry using geometric predicates. The planner operates on $x_t$, while the world model evolves $z_t$; the mapping $\mathcal{P}$ serves as the interface between perception and planning.

\subsection{Action-Conditioned World Model Dynamics}
\setlength{\intextsep}{4pt}
\setlength{\columnsep}{10pt}
\begin{wrapfigure}{R}{0.48\textwidth} 
  \vspace{-38.5pt}
  \centering
  \includegraphics[trim={2.0cm, 1cm, 1cm, 1cm},width=0.48\textwidth]{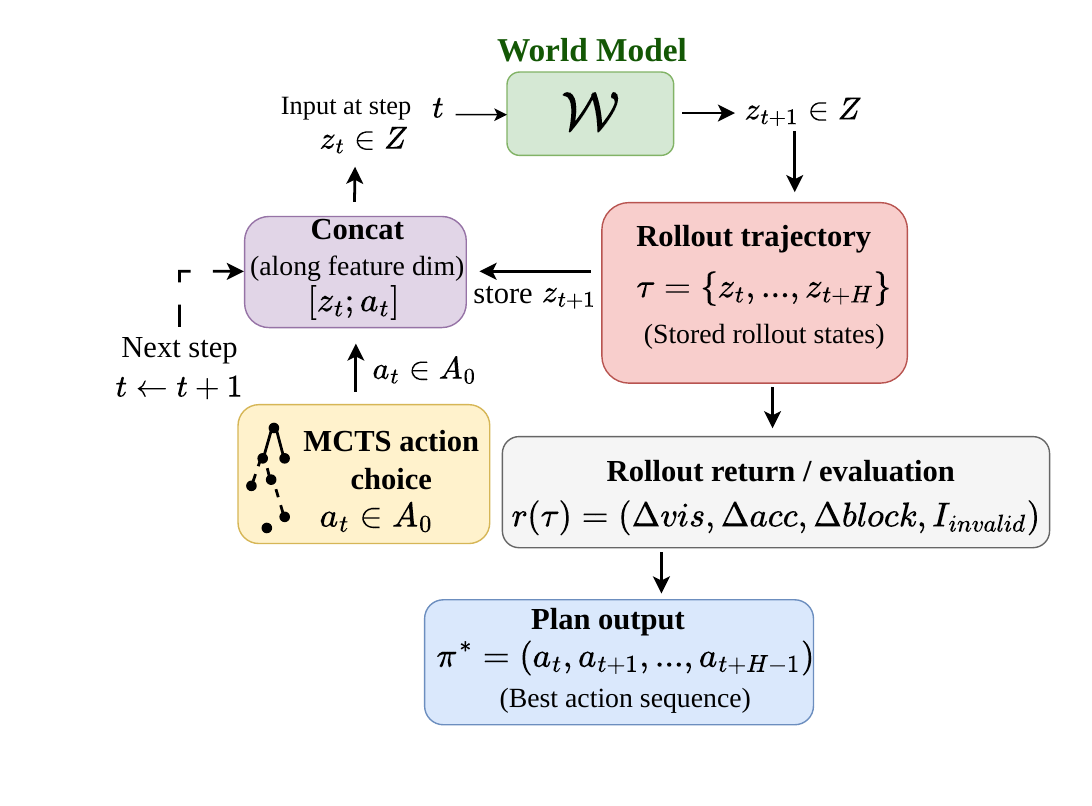}
  \caption{\textbf{WM-Guided Planning:} Starting from the reconstructed scene state, future states are recursively predicted by the action-conditioned world model during MCTS rollouts, and the resulting summaries guide final action selection.}
  \vspace{+11.8pt}
  \label{fig:wm_generation}
\end{wrapfigure}

We learn a dynamics model $z_{t+1} = f_{\phi}(z_t, a_t)$. The model predicts per-object updates $\{(\Delta \mu_i, m_i, v_i)\}_{i=1}^K$, where $\Delta \mu_i \in \mathbb{R}^3$ is the displacement, $m_i \in \{0,1\}$ indicates whether object $i$ moves, and $v_i \in \{0,1\}$ denotes whether the predicted update is physically valid. The next state is obtained by $\mu_i^{t+1} = \mu_i^t + m_i \cdot \Delta \mu_i$, while $\sigma_i$ and $p_i$ remain unchanged. This abstraction targets rigid-object, no-add/remove primitives;
orientation is not explicitly represented, and visibility is
recomputed from geometry. The world model is trained in two phases: an initial phase on simulated object trajectories, followed by a fine-tuning phase on \ghost-reconstructed scene representations. During fine-tuning, hallucinated slots (those without a matched ground-truth object) are supervised to produce zero displacement, improving robustness to reconstruction noise.

\subsection{LLM-Guided Action Proposals}

We use the LLM to generate a structured set of action proposals from the task goal $g$ and initial planning state $x_0$. The LLM is instructed to read $g$ and $x_0$, then propose task-relevant object subsets and coarse motion directions; feasibility and final plan selection are left to the world model and MCTS. Each proposal is parameterized as $c = (\mathcal{I}_c, d_c)$, where $\mathcal{I}_c \subseteq \{1,\dots,K\}$ is a subset of object indices and $d_c \in \{\texttt{left}, \texttt{right}, \texttt{back}, \texttt{forward}\}$ is a motion direction. Each proposal $c=(\mathcal{I}_c,d_c)$ restricts search to a smaller candidate action set. Here, $\mathcal{I}_c$ specifies the subset of objects to consider, and $d_c$ specifies the preferred motion direction. We therefore construct
\begin{equation}
\mathcal{A}(c) = \{(i,d_c) \mid i \in \mathcal{I}_c\} \cup \mathcal{A}_{\text{aux}},
\end{equation}
where $\mathcal{A}_{\text{aux}}$ contains a small set of fallback actions generated at each root from geometric predicates in the current relational state, such as pulling the target-facing blocker forward or shifting the nearest corridor blocker laterally. These auxiliary actions prevent the proposal set from becoming overly narrow when the LLM omits a useful object or direction. If multiple proposals are generated, we take the union of their candidate sets. MCTS then searches over this reduced action space instead of all object-primitive combinations.

\subsection{Grounded Selection and Search}
\begin{wrapfigure}{r}{0.48\textwidth}
\vspace{-32pt}
\begin{minipage}{0.48\textwidth}
\begin{algorithm}[H]
\small
\caption{SCOUT Planning}
\label{alg:scout}
\begin{algorithmic}[1]
\Require observations $o$, task goal $g$, planning horizon $H$,
         learned dynamics $f_\phi$, action budget $T$
\Ensure Selected action sequence $\pi^*$

\State $z_0 \leftarrow \ghost(o)$; $\pi^* \leftarrow [\,]$
\State $x_0 \leftarrow \mathcal{P}(z_0, g)$
\State $\mathcal{C} \leftarrow \text{LLM}(x_0)$

\While{$|\pi^*| < T$ and $\neg\mathrm{Success}(x_0,g)$}
    \State $\mathcal{A}_0 \leftarrow
           \bigcup_{c \in \mathcal{C}} \mathcal{A}(c)$
    \State Initialize search tree $\mathcal{T}$ with root node $z_0$
    \While{within search budget}
        \State $\tau \leftarrow
               \text{SelectAndExpand}(\mathcal{T}, \mathcal{A}_0)$
        \For{$t = 0$ to $H-1$}
            \State $z_{t+1} \leftarrow f_\phi(z_t, a_t)$
        \EndFor
        \State $r(\tau) \leftarrow \text{Summarize}(\tau)$
        \State $\text{Backpropagate}(r(\tau), \mathcal{T})$
    \EndWhile

    \State $\hat{\pi} \leftarrow
           \argmax_{\mathbf{a}_{1:H} \in \mathcal{T}}
           \mathbb{E}[\mathrm{Success}(z_H, g)]$
    \State $a \leftarrow \hat{\pi}_1$; execute $a$;
           append $a$ to $\pi^*$
    \State $z_0 \leftarrow f_\phi(z_0,a)$
    \State $x_0 \leftarrow \mathcal{P}(z_0,g)$
\EndWhile

\State \Return $\pi^*$



\end{algorithmic}
\end{algorithm}
\end{minipage}
\vspace{-1pt}
\end{wrapfigure}
Given the candidate action set, we simulate future trajectories $\tau$ using the world model, as illustrated in Fig.~\ref{fig:wm_generation}. Each rollout is summarized by $r(\tau) = \big(\Delta \mathrm{vis}(\tau),\; \Delta \mathrm{acc}(\tau),\; \Delta \mathrm{block}(\tau),\; \mathbb{I}_{\mathrm{invalid}}(\tau)\big)$,
where $\Delta \mathrm{vis}$ measures visibility gain, $\Delta \mathrm{acc}$ measures accessibility progress, and $\mathbb{I}_{\mathrm{invalid}}$ flags physically invalid transitions. These summaries are produced by an MCTS procedure that constructs a search tree over predicted scene states using the learned world model. At each decision, MCTS evaluates depth-$H=2$ rollouts and commits only the first primitive of the highest-scoring branch. The root state is updated and the tree rebuilt until predicted success or the $T=12$ action budget, enabling long-horizon task completion through receding-horizon planning. The initial LLM proposals are retained throughout. In Algorithm~\ref{alg:scout}, rollout indices are local to each search, $z_0$ and $x_0$ denote the current root, and $\mathrm{Success}(x_0,g)$ denotes predicted success. In the closed-loop variant, new RGB-D observations refresh $z_0$ at the specified sensing interval before recomputing $x_0$.

\section{Experiments}

We conduct a comprehensive set of experiments to evaluate the effectiveness, generality, and robustness of \method for manipulation under occlusion. Our evaluation is organized around four questions: 1) Does \method improve task success in cluttered, partially observable manipulation? 2) How well does it generalize to unseen objects, view layouts, and harder occlusion settings? 3) Which components of the planner contribute most to performance? 4) Can our method successfully transfer to the real world?

\begin{figure}[t]
    \centering
    \includegraphics[trim={1cm 8.1cm 1cm 1cm}, width=\textwidth]{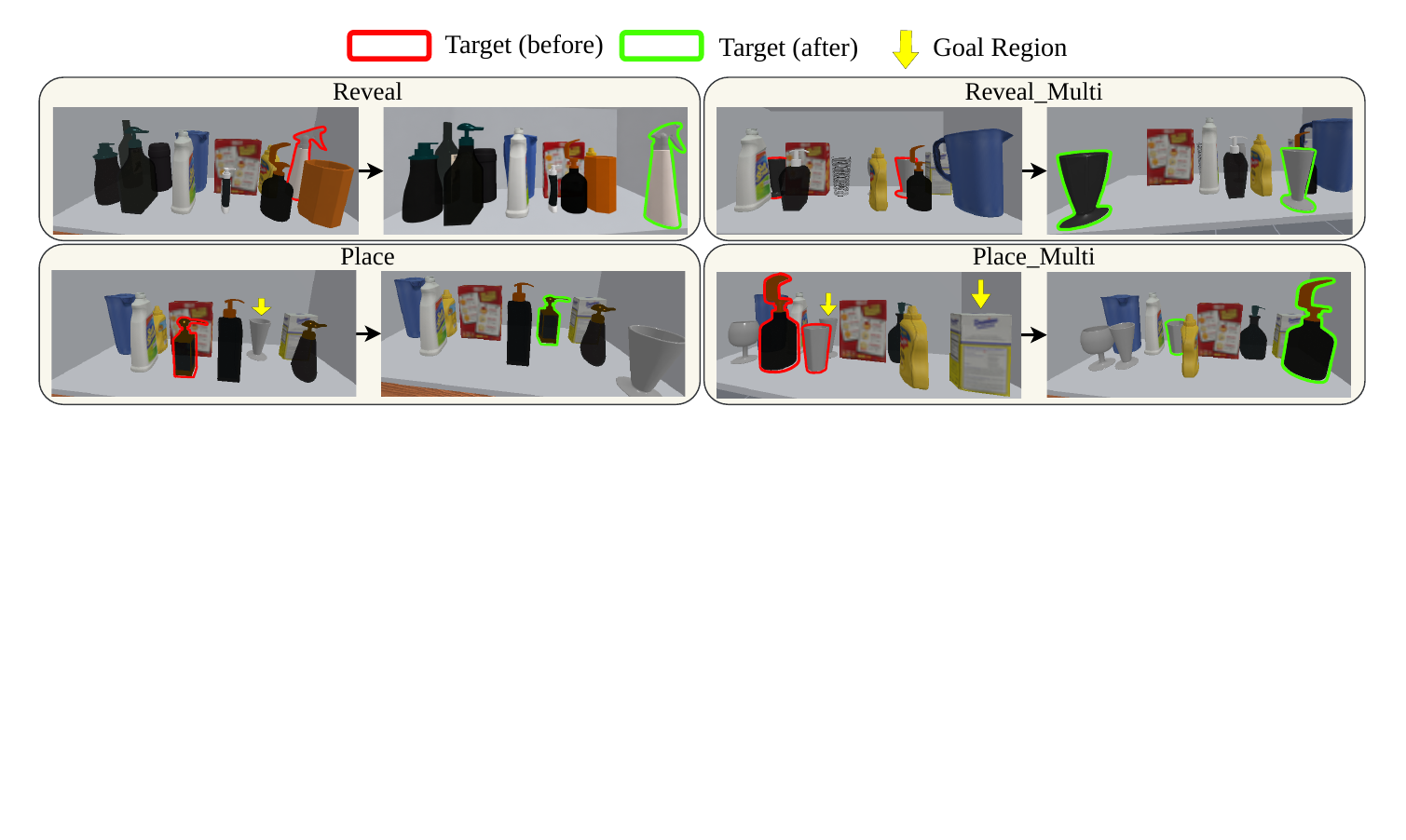}
    \caption{Representative examples from the four task families used for evaluation.}
    \label{fig:benchmarks}
\end{figure}

\paragraph{Task suite.}
We evaluate \method on a cluttered shelf benchmark designed to isolate the hidden-scene planning problem under controlled occlusion. Scenes are generated in ManiSkill2~\citep{gu2023maniskill2} using a procedurally constructed shelf and object assets from the ManiSkill2-YCB set~\cite{calli2017ycb}, ProcTHOR~\cite{delitzas2023procthor}, and RoboCasa~\cite{nasiriany2024robocasa}. Each scene contains 8--16 objects, including task objects, foreground blockers, and distractors; object poses are sampled so that the target object or placement region is initially occluded from the frontal view. In the initial-view protocol, each scene is observed from a small set of RGB-D viewpoints sampled around the shelf opening, and the planner must produce a finite-horizon sequence of object-centric primitives from these sparse observations alone.
We evaluate the four task families introduced in Sec.~3.1 and illustrated in Fig.~\ref{fig:benchmarks}: \textbf{Reveal}, \textbf{Reveal\_Multi}, \textbf{Place}, and \textbf{Place\_Multi}. 
Additional details on shelf generation, object assets, camera sampling, and task construction are provided in Section~\ref{app:data_generation}.

\paragraph{Implementation Details.}
We use Qwen2.5-7B-Instruct~\cite{yang2024qwen25} as the open-source action-proposal model in \method. We set the rollout scoring weights to $\alpha_{\mathrm{vis}}=1.0$, $\alpha_{\mathrm{acc}}=1.0$, $\alpha_{\mathrm{block}}=0.25$, and $\alpha_{\mathrm{invalid}}=1.0$. All learned components are trained with AdamW~\cite{loshchilov2019decoupled} on simulator-generated shelf data. Experiments are conducted on four NVIDIA L40S-48GB GPUs with 64 CPU cores and 512GB RAM; full architecture, hyperparameter, compute, and inference details are provided in Section~\ref{app:implementation}.

\paragraph{Baselines.}
For the initial-view comparison, we adapt four reference methods to the same sparse-observation, no-reobservation, and discrete-primitive setting. The VLA comparisons characterize performance under this shared interface rather than native continuous-control performance.
\begin{itemize}
    \item $\pi_{0.5}$ \citep{black2025pi05} is an end-to-end vision-language-action policy designed for open-world robot manipulation, mapping visual observations and task specifications to continuous low-level robot actions. For fair comparison, we provide the same initial RGB-D views and a templated task specification, decode each predicted action to the nearest valid primitive and execute the resulting primitive sequence under the same no-reobservation protocol as \method.
    \item Long-VLA \citep{fan2025longvla} is a long-horizon VLA policy designed to extend vision-language-action models to multi-step manipulation. Since it originally produces continuous robot actions, we evaluate it with the same initial observation set and task specification as \method, map its outputs to the shared primitive vocabulary, and impose the same $T=12$ action budget.
    \item OWMM-Agent \citep{chen2025owmmagent} is an embodied agent that combines scene understanding, state tracking, and action generation for open-world mobile manipulation. To adapt it to our stationary shelf benchmark, we disable navigation and auxiliary tool use, provide only the same initial RGB-D observations, and require all decisions to be emitted through the same object-centric primitive API.
    \item ParticleFormer \citep{huang2025particleformer} is a 3D point-cloud world model for predicting action-conditioned scene evolution rather than a complete high-level planner. We therefore use it as a drop-in replacement for \method's learned transition model, while keeping the reconstructed planning state, candidate action set, rollout objective, and search budget fixed.
\end{itemize}

\paragraph{Evaluation Metrics.}
We evaluate performance using task-completion success rates over $500$ independent rollouts with randomly initialized shelf configurations. Since \method outputs high-level manipulation primitives rather than low-level controls, we measure trajectory length in primitive actions and impose a maximum budget of $T=12$ actions per episode. Reported success is computed from simulator ground-truth geometry: Reveal rollouts are counted as successful if the target object's visibility and accessibility scores both exceed $0.6$ within $T$ actions, while Place rollouts are successful if the specified object reaches its assigned goal region within $T$ actions. During planning, these quantities are estimated from the relational planning state $x_t=P(z_t,g)$ for rollout scoring. In addition to success rate, we report the mean number of primitive actions per successful episode, target visibility gain, target accessibility gain, and invalid-action rate. The visibility score is computed from a fixed front-facing evaluation camera. For Reveal tasks, it measures how much of the target object is visible; for Place tasks, it measures how much of the goal region is visible, normalized by its empty-shelf projection. The accessibility score measures whether the task site is reachable from the shelf opening: it is high when the path to the target object or goal region is clear, and low when other objects block this path. Actions are marked invalid when the predicted transition violates geometric feasibility constraints. Metric construction details are provided in Section~\ref{app:planning_state_metrics}.

\subsection{Simulation Results}

\begin{figure}[t]
    \centering
    \includegraphics[width=\textwidth]{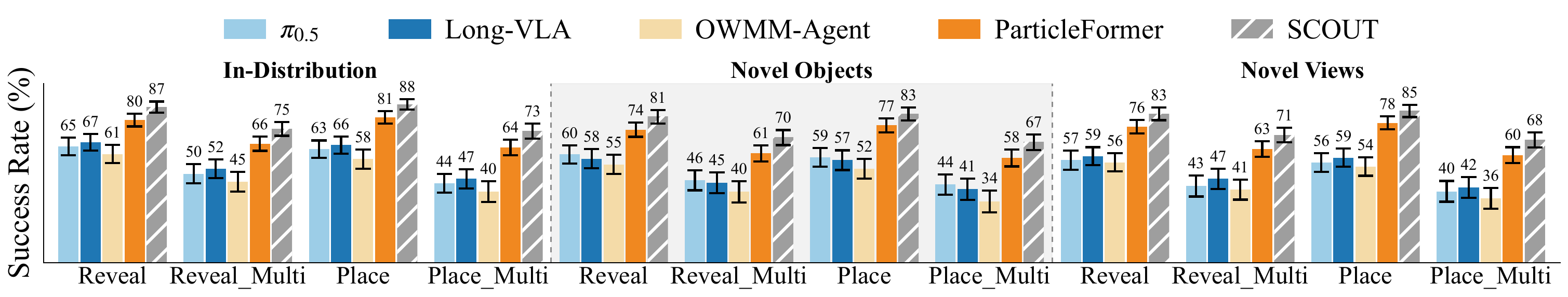}
    \caption{
Results on the simulated shelf task suite. Success rate (in \%) is averaged across $3$ seeds and uncertainty is given by the standard error. \method consistently outperforms the adapted baselines across in-distribution, novel-object, and novel-view settings.
}
    \label{fig:results}
\end{figure}

Figure~\ref{fig:results} shows the success rates for the simulation tasks under in-distribution, novel-object, and novel-view settings. \method achieves the highest success rate across all task families and distributions, reaching an in-distribution average success of $80.8\%$ compared with $72.8\%$ for the strongest baseline. The gains are largest in the multi-object settings, where \method improves over the strongest baseline by $9.0$ points on Reveal\_Multi and $9.0$ points on Place\_Multi. These settings require reasoning over hidden blockers, intermediate rearrangements, and narrow access corridors. The same trend holds in the single-object tasks, where \method reaches $86.8\%$ on Reveal and $88.2\%$ on Place, improving over the strongest baseline by $7.2$ points in each case. These results use the initial-view protocol; Appendix~\ref{app:simulation_eval} reports closed-loop comparisons and additional proposal, search, calibration, and reconstruction diagnostics.

\textbf{Generalization Performance.} \method remains strong under distribution shift. On novel objects, it reaches an average success of $75.4\%$, compared with $67.6\%$ for the strongest baseline. On novel views, it reaches $76.8\%$, compared with $69.3\%$. The smaller degradation under viewpoint and object shifts suggests that planning over reconstructed object-centric structure provides a stable basis for manipulation under occlusion.

\newcommand{\smallstd}[1]{\, \text{\scriptsize $\pm$#1}}

\begin{wraptable}{r}{0.48\textwidth}
    \centering
    \vspace{-8pt}
    \scriptsize 
    \setlength{\tabcolsep}{2pt} 
    \caption{Trajectory rollout metrics.}
    \label{tab:traj_metrics}
    \begin{tabular}{lcccc}
        \toprule
        Task & Steps & $\Delta$ Visibility & $\Delta$ Access. & Invalid Rate \\
        \midrule
        Reveal       & 6.3{\tiny$\pm$1.4}  & +0.46{\tiny$\pm$0.07} & +0.41{\tiny$\pm$0.06} & 3.8\% \\
        Reveal\_Multi    & 8.7{\tiny$\pm$1.8}  & +0.59{\tiny$\pm$0.09} & +0.53{\tiny$\pm$0.08} & 5.9\% \\
        Place        & 6.0{\tiny$\pm$1.2}  & +0.34{\tiny$\pm$0.06} & +0.37{\tiny$\pm$0.05} & 3.2\% \\
        Place\_Multi    & 9.1{\tiny$\pm$2.0}  & +0.51{\tiny$\pm$0.08} & +0.49{\tiny$\pm$0.07} & 6.4\% \\
        \bottomrule
    \end{tabular}
    \vspace{-4pt}
\end{wraptable}

\textbf{Planning Efficiency.} The plan-level diagnostics in Table~\ref{tab:traj_metrics} show that successful SCOUT rollouts are shorter in the single-object settings and longer in the multi-object settings. Reveal and Place require $6.3$ and $6.0$ primitive actions on average, whereas Reveal\_Multi and Place\_Multi require $8.7$ and $9.1$. The multi-object tasks also produce larger visibility and accessibility gains, reflecting that more occlusion and access constraints must be cleared before the target or goal region becomes usable. Invalid-action rates remain low across all four families, ranging from $3.2\%$ to $6.4\%$, indicating that search rarely selects geometrically infeasible primitives.

\subsection{Failure Analysis and Qualitative Results}

\begin{figure}[t]
    \centering
    \includegraphics[width=\textwidth]{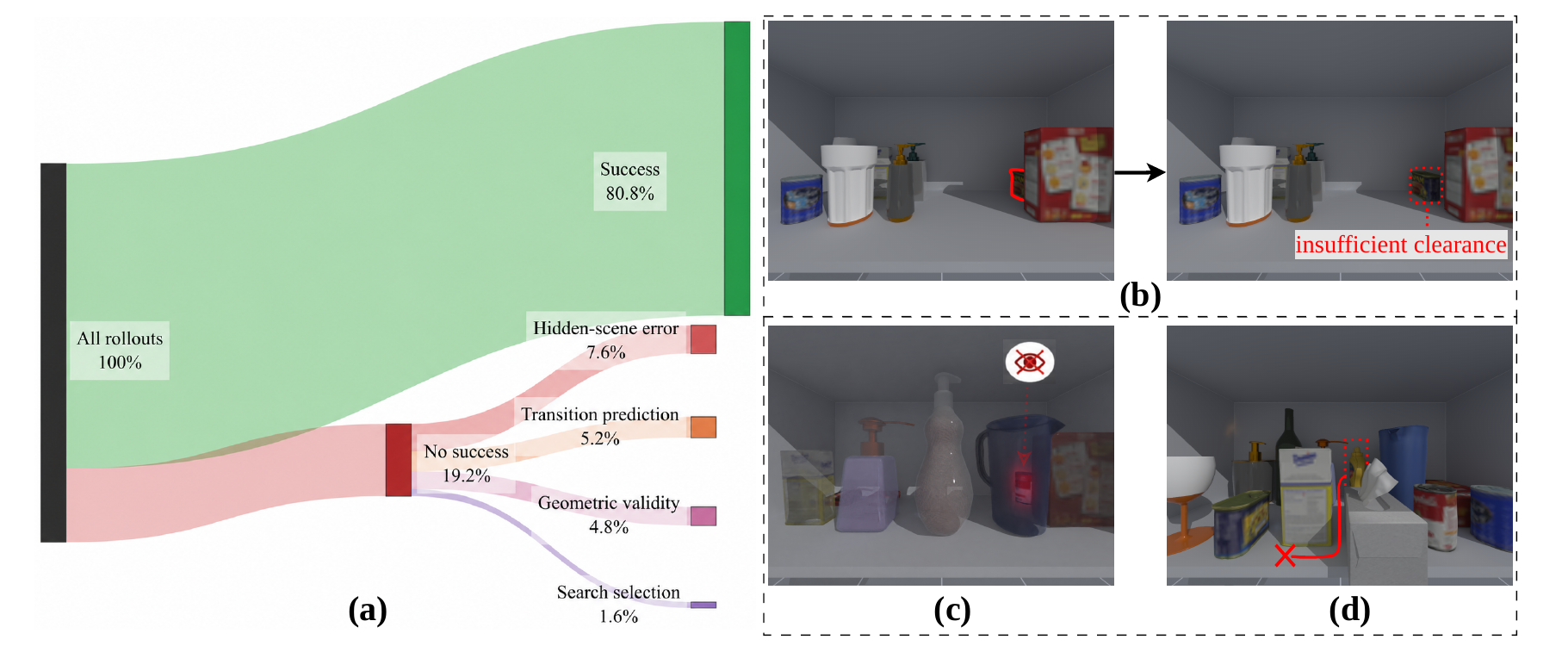}
    \caption{
\textbf{Failure analysis of \method.} 
\textbf{(a)} Failure decomposition as percentages of all rollouts. Most failures stem from hidden-scene reconstruction errors, followed by transition-prediction and geometric-validity errors. 
\textbf{(b)} Transition-prediction failure: the selected action does not create enough free space to retrieve the target. 
\textbf{(c)} Hidden-scene failure: a small back-row target remains too occluded to be reconstructed from partial observations. 
\textbf{(d)} Geometric-validity failure: dense clutter leaves no feasible retrieval corridor under the current primitive set.
}
    \vspace{-0.5cm}\label{fig:failure}
\end{figure}

\textbf{Failure Modes.} To analyze where unsuccessful rollouts break down, we categorize failures in Fig.~\ref{fig:failure}a into four sources: hidden-scene errors, transition-prediction errors, geometric-validity errors, and search-selection errors. The largest source is hidden-scene error, accounting for $7.6\%$ of all rollouts, where the reconstructed state misses a blocker or overestimates the free corridor to the task site. 
Transition-prediction errors account for $5.2\%$, typically when the rollout predicts that an action will clear visibility or access but the executed transition leaves the site partially blocked. Geometric-validity failures contribute $4.8\%$, while search-selection errors are the smallest category at $1.6\%$, indicating that most failures arise before final action selection rather than from the scoring rule itself.

Panels (b)--(d) show representative failure cases behind the decomposition in Fig.~\ref{fig:failure}. In (b), the world model predicts that the selected action will create enough clearance, but execution leaves the target only partially exposed and still not retrievable. In (c), the target is too small and occluded in the back row, so it is not recovered by the hidden-scene reconstruction. In (d), the scene is so congested that the available shift and pull primitives cannot open a feasible access corridor without an explicit remove action.

\setlength{\intextsep}{4pt}
\setlength{\columnsep}{10pt}
\begin{wrapfigure}{R}{0.48\textwidth} 
  \vspace{-1.5pt}
  \centering
  \includegraphics[trim={0.5cm, 0.5cm, 1cm, 1cm},width=0.48\textwidth]{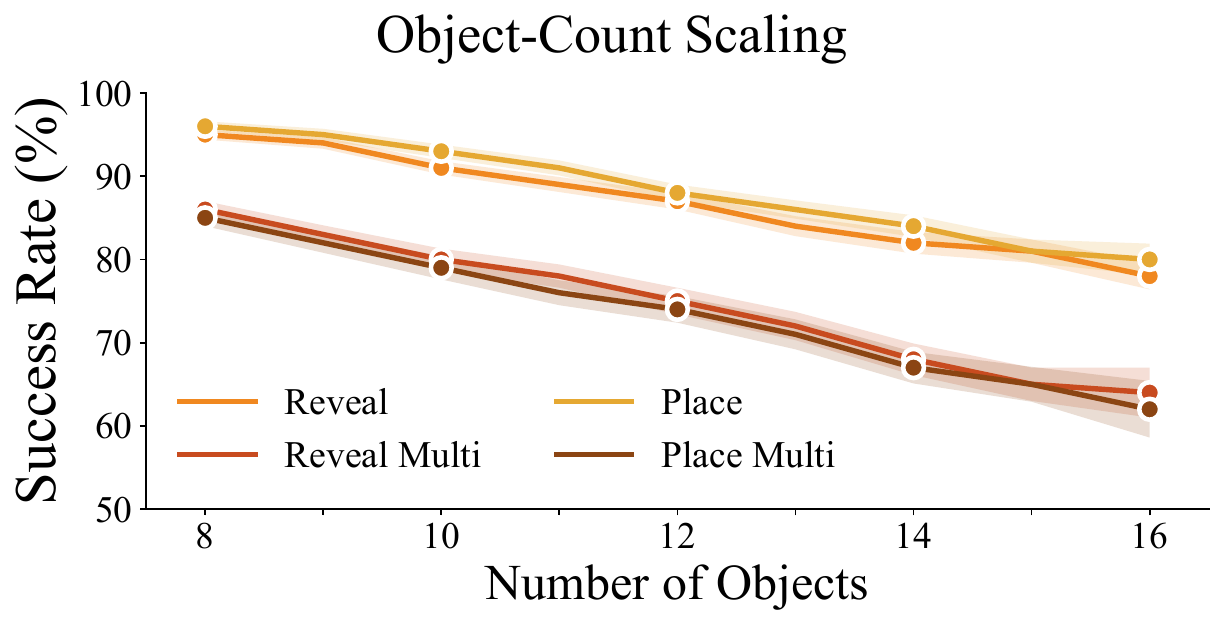}
  \caption{Success rate as clutter density increases.}
  \vspace{+1.8pt}
  \label{fig:object_count_scaling}
\end{wrapfigure}
\textbf{Scaling with Clutter Density.}
In Fig.~\ref{fig:object_count_scaling}, we evaluate how task success changes as the number of shelf objects increases from $8$ to $16$. Performance decreases gradually across all task families, showing that denser clutter makes occlusion reasoning and rearrangement more difficult. The drop is mild for single-object Reveal and Place tasks, while Reveal\_Multi and Place\_Multi degrade more sharply as additional objects introduce more blockers and longer rearrangement chains. Even at $16$ objects, \method maintains reasonable success, indicating that the planner remains effective under high clutter.

\begin{wraptable}{r}{0.48\textwidth}
    \centering
    \vspace{-8pt}
    \scriptsize 
    \setlength{\tabcolsep}{2pt} 
    \caption{Ablation study of \method components.}
    \label{tab:ablation}
        \begin{tabular}{lcc}
            \toprule
            Variant & Avg. Succ. (\%) & Avg. Act. \\
            \midrule
            Single-view input         & 69.7{\tiny$\pm$4.6} & 7.9{\tiny$\pm$0.5} \\
            w/o relational projection & 68.1{\tiny$\pm$4.4} & 8.2{\tiny$\pm$0.5} \\
            w/o rollout search        & 70.6{\tiny$\pm$4.2} & 6.8{\tiny$\pm$0.4} \\
            \textbf{Full \method}     & 80.8{\tiny$\pm$3.5} & 7.5{\tiny$\pm$0.4} \\
            \bottomrule
        \end{tabular}
    \vspace{-4pt}
\end{wraptable}
\textbf{Ablation Study.}
Table~\ref{tab:ablation} shows that each component contributes to final performance. Using a single input view reduces average success to $69.7\%$, indicating that multi-view observations improve the reconstructed scene state under occlusion. Removing the relational projection gives the lowest success, $68.1\%$, and also increases the average plan length to $8.2$ actions, showing that explicit task-site, blocker, and spatial relations are important for efficient search. Replacing rollout search with greedy depth-1 selection reaches $70.6\%$ success and uses fewer actions on average, but the shorter plans are less reliable because they do not evaluate longer rearrangement effects. The full model achieves the best success, $80.8\%$, while maintaining a moderate action length of $7.5$ primitives.

\subsection{Real-World Experiments}
\begin{figure}[t]
    \centering
    \includegraphics[width=\textwidth]{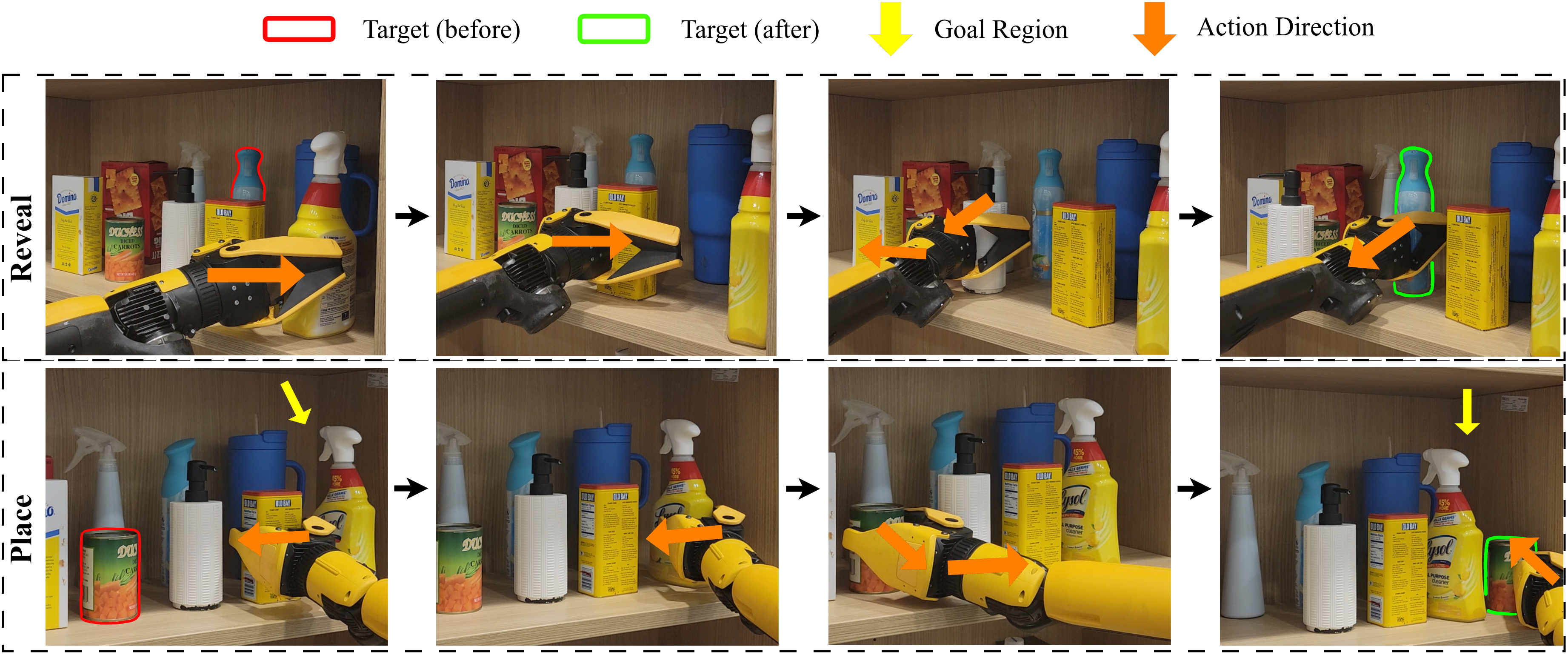}
    \caption{Real-world rollouts of \method on the \textsc{Reveal} and \textsc{Place} tasks in cluttered shelf scenes.}
    \vspace{-0.5cm}\label{fig:demo}
\end{figure}
To evaluate real-world performance, we deploy \method on a Boston Dynamics Spot platform~\cite{spot_arm_specs} equipped with a 6-DOF manipulator and its integrated gripper-mounted 4K RGB and Time-of-Flight (ToF) depth sensors. Observations are captured from static frontal viewpoints to maintain simulation-consistent partial visibility, with action candidates executed via a kinematic motion controller. We perform 10 trials each for the \textbf{Reveal} and \textbf{Place} tasks using a variety of household objects on a cluttered shelf. 

Across these 20 trials, \method achieves a \textbf{60\% (12/20)} success rate. As illustrated in Fig.~\ref{fig:demo}, the system successfully reasons over hidden topology to plan strategic rearrangements, such as shifting a blocking container (\textbf{orange arrows}) to expose an occluded target (\textbf{red to green highlight}) or clearing an access corridor to reach a goal region (\textbf{yellow arrow}). Failures were distributed between low-level hardware issues, such as gripper slippage on compliant surfaces, and plan-level errors. The latter primarily involved the world model proposing physically invalid actions or MCTS rollouts failing due to noisy \ghost reconstructions in high-glare environments. Detailed hardware specifications and individual failure case analysis in Section ~\ref{app:real_robot}.

\section{Conclusion and Limitations}

In this work, we presented SCOUT, a planning framework for manipulation under partial observability that integrates hidden-scene reconstruction, object-centric action abstraction, and world-model-guided rollout evaluation. SCOUT enables structured reasoning over occluded environments by combining LLM-guided action proposals with grounded evaluation of action-conditioned futures. We show that this approach improves task success across cluttered shelf manipulation settings, including multi-step rearrangement and generalization to unseen configurations. The framework operates from sparse RGB-D observations and a discrete primitive interface, supporting consistent behavior across simulation and real-world deployment.

\textbf{Broader Impact.} SCOUT improves the ability of robotic systems to operate in cluttered and partially observable environments, with potential applications in warehouse automation, assistive robotics, and service tasks. By reasoning about hidden objects and access corridors, it can reduce manipulation errors and improve reliability in real-world settings. The structured planning interface may also support more interpretable decision-making in human-robot interaction scenarios.

\textbf{Limitations.} Several limitations remain. The performance of SCOUT depends on the accuracy of hidden-scene reconstruction, and errors in estimated geometry, object existence, or blocker relationships can propagate through planning. The learned world model may produce inaccurate predictions under complex contact interactions or out-of-distribution object arrangements. In addition, the current formulation assumes a static scene during planning and relies on a fixed set of discrete manipulation primitives, which may limit applicability to dynamic environments or tasks requiring continuous contact-rich control.

\begin{ack}
This work has taken place in the RobotiXX Laboratory at George Mason University. RobotiXX research is supported by National Science Foundation (NSF, 2350352), Army Research Office (ARO, W911NF2320004, W911NF2520011), Army Ground Vehicle Systems Center (GVSC), Google DeepMind (GDM), Microsoft Research (MSR), Clearpath Robotics, FrodoBots Lab, Raytheon Technologies (RTX), Tangenta, 4-VA, Mason Innovation Exchange (MIX), and Walmart. The authors declare no competing interests.
\end{ack}

\bibliographystyle{plainnat}
\bibliography{references}

\newpage
\appendix

\section{Implementation Details}
\label{app:implementation}

This appendix provides additional implementation details for \method, complementing the method and experimental setup in the main paper. We describe the simulator, data generation, scene reconstruction module, relational state construction, world-model training, MCTS rollout evaluation, and real-robot deployment.

\subsection{Simulator and Dataset Generation}
\label{app:data_generation}

\paragraph{Shelf environment.}
All simulated scenes are generated in ManiSkill2 with SAPIEN physics~\cite{gu2023maniskill2}. We use a procedurally instantiated cluttered shelf with approximate dimensions $0.36\,\mathrm{m}\times0.90\,\mathrm{m}\times0.40\,\mathrm{m}$ and a fixed shelf-frame convention: depth points from the shelf opening toward the back wall, the lateral axis spans left--right, and the vertical axis points upward. Candidate object placements are sampled from a $2\times6$ slot grid inside the shelf volume.

\paragraph{Object assets.}
We construct cluttered shelf scenes from ManiSkill2-YCB objects~\cite{calli2017ycb}, ProcTHOR / AI2-THOR assets~\cite{delitzas2023procthor}, and RoboCasa kitchen assets~\cite{nasiriany2024robocasa}. YCB provides everyday graspable objects such as cans, boxes, mugs, and bottles, while ProcTHOR and RoboCasa provide additional household and kitchen objects. Objects are sampled from shelf-scale categories including mugs, cans, boxes, bottles, and jars. For stable simulation, visual meshes are rendered from the original assets, while collision geometry is approximated by box collision shapes. We use $77$ YCB objects, $97$ ProcTHOR objects, and $16$ RoboCasa objects.

\paragraph{Scene sampling.}
Each scene is sampled by choosing a difficulty level and populating the shelf with objects from the asset pool. Easy, medium, and hard scenes contain $4$--$8$, $7$--$12$, and $10$--$16$ objects, respectively; the benchmark settings in the main paper use $8$--$16$ objects. Each scene contains task objects, foreground blockers, and distractors. Distractors are objects that are not directly part of the task but may still affect planning if they block the access corridor. Single-object tasks contain one task object, while multi-object tasks contain two task objects. Each scene contains $2$--$5$ foreground blockers, with the remaining objects treated as distractors. Candidate poses are sampled with jitter and rejected if they violate shelf bounds, intersect other objects, or fail the initial occlusion requirement.

\paragraph{Task families.}
As described in Sec.~\ref{sec:problem_formulatio} and illustrated in Fig.~\ref{fig:benchmarks}, we evaluate four task families: \textbf{Reveal}, \textbf{Reveal\_Multi}, \textbf{Place}, and \textbf{Place\_Multi}. Reveal specifies one occluded target object, while Reveal\_Multi specifies two targets. Place specifies one object and an axis-aligned goal region, while Place\_Multi specifies two objects and a shared goal region. In the reported experiments, the goal region occupies the front $30\%$ of the shelf depth, spans the valid lateral workspace, and uses the lower half of the shelf height as the placement volume. All tasks use a maximum episode budget of $T=12$ primitive actions.

\begin{table}[ht]
    \centering
    \small
    \caption{Task construction details for the simulated shelf benchmark.}
    \label{tab:app_task_families}
    \begin{tabular}{l c l}
        \toprule
        Task & Targets & Success condition \\
        \midrule
        Reveal & 1 & Target visibility and accessibility exceed threshold \\
        Reveal\_Multi & 2 & Both targets satisfy Reveal predicate \\
        Place & 1 & Target object lies inside assigned goal region \\
        Place\_Multi & 2 & Both target objects lie inside the shared goal region \\
        \bottomrule
    \end{tabular}
\end{table}

\paragraph{Camera sampling.}
Each scene is observed from sparse front-facing RGB-D viewpoints, matching the partial-observation setting in Fig.~\ref{fig:GHOST}. We generate $16$ candidate views by varying camera yaw, pitch, lateral offset, height offset, and distance from the shelf opening, and select up to $5$ views per episode. The RGB-D resolution is $896\times504$. Depth is represented in metres, clamped to $[0.05,3.0]$, and back-projected using the camera intrinsics. RGB images are normalized with ImageNet mean and standard deviation. At this resolution, we use the approximate intrinsics
\[
K =
\begin{bmatrix}
640.0 & 0 & 448.0 \\
0 & 640.0 & 252.0 \\
0 & 0 & 1
\end{bmatrix}.
\]

\begin{table}[ht]
    \centering
    \small
    \caption{Camera sampling ranges used for simulated RGB-D observations.}
    \label{tab:app_camera_sampling}
    \begin{tabular}{l c}
        \toprule
        Parameter & Values / Range \\
        \midrule
        Candidate views & $16$ \\
        Selected views & up to $5$ \\
        Distance from shelf opening & $0.34$--$0.46$ m \\
        Yaw & $\{-20,-12,0,12,20\}^{\circ}$ \\
        Pitch & $\{0,-8\}^{\circ}$ \\
        Lateral offset & $\{-0.05,0.0,0.05\}$ m \\
        Height offset & $\{-0.08,0.0,0.08\}$ m \\
        RGB-D resolution & $896\times504$ \\
        Depth range & $[0.05,3.0]$ m \\
        \bottomrule
    \end{tabular}
\end{table}

\paragraph{Transition data for world-model training.}
For each simulated state, we store oracle object poses, object sizes, relational labels, RGB-D observations, segmentation masks, occupancy volumes, object-slot ground truth, and reconstructed slot states. To construct the transition dataset, we sample a valid cluttered scene, record the pre-action state, execute an object-centric primitive $(i,u)$ in simulation, allow physics to settle, and record the post-action state. Transition labels are obtained by comparing the pre- and post-action object states. The displacement target is $\Delta\mu_i=\mu_i^{t+1}-\mu_i^t$, the moved-object mask marks objects whose centers move by more than $5\times10^{-4}$ m, and the validity label indicates whether the transition remains geometrically feasible within the shelf bounds. Invalid transitions include no-motion actions, missing-object actions, severe collisions, and motions that move objects outside the valid workspace.

Primitive actions for transition generation are sampled by a heuristic back-to-front policy that prioritizes front-row blockers and actions likely to expose deeper objects, while still sampling lateral shifts and forward pulls to cover diverse local interactions. The reported training set contains $2000$ simulated scenes and the validation set contains $500$ simulated scenes. Each scene contributes up to $12$ primitive transitions, yielding approximately $24{,}000$ training transitions and $6{,}000$ validation transitions before filtering invalid or duplicate transitions.

\paragraph{Evaluation splits.}
Each simulation setting in Fig.~\ref{fig:results} is evaluated over $500$ rollouts and $3$ random seeds. The in-distribution split uses the same asset pool, camera ranges, and object-count distribution as training. The novel-object split holds out $20\%$ of assets from each source and samples task objects from the held-out pool. The novel-view split uses yaw angles $\{\pm28,\pm35\}^{\circ}$, camera distances $0.48$--$0.56$ m, and height offsets up to $\pm0.12$ m. For clutter-density evaluation, we use object counts $\{8,10,12,14,16\}$.

\subsection{GHOST Scene Reconstruction}
\label{app:ghost_details}
\begin{figure}[ht]
    \centering
    \includegraphics[width=0.9\textwidth]{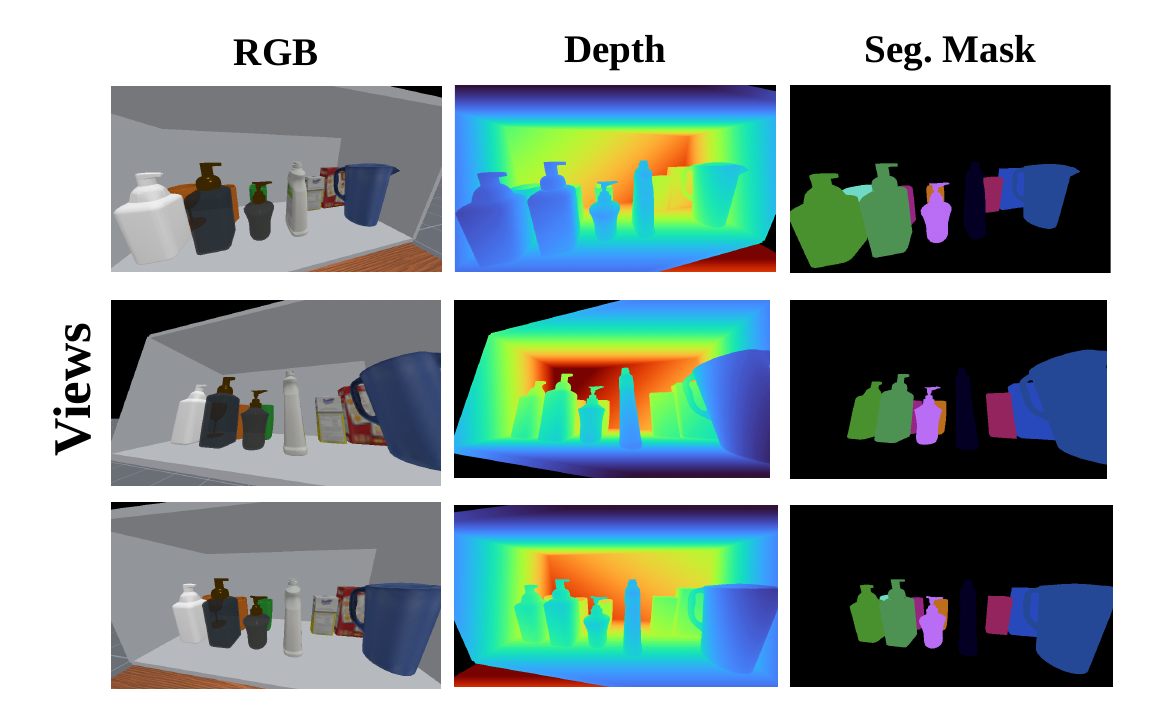}
    \caption{Example sparse RGB-D observations used by GHOST. Each row shows a different front-facing view of the same cluttered shelf scene, with corresponding RGB image, depth map, and object segmentation mask.}
\label{fig:app_ghost_inputs}
\end{figure}

\paragraph{Input representation.}
GHOST receives RGB-D observations, camera intrinsics, camera extrinsics, and object-level priors. For each view, the input consists of an RGB image, a depth map in metres, an intrinsic matrix $K$, a world-from-camera transform $T_{wc}$, and object masks. In simulation, object masks are obtained from simulator segmentation when available; otherwise SAM3D is used as an object-level prior~\cite{sam3d2025}. The planner uses up to $5$ views from the candidate view set described above. Fig.~\ref{fig:app_ghost_inputs} shows representative RGB-D observations and segmentation masks used as input to GHOST before cross-view fusion.

\paragraph{Object priors and cross-view fusion.}
For each RGB-D view, SAM3D provides an object mask, coarse 3D geometry from mask and depth, a confidence score, and a learned shape token. We use a confidence threshold of $0.5$ and shape-token dimension $64$. Per-view hypotheses are associated in 3D using coarse object boxes. Detections with 3D-IoU above $0.25$ are merged into a single visible-slot initialization. Object centers and sizes are averaged using confidence weights, existence probability is set to the maximum confidence in the cluster, and shape tokens are aggregated by confidence-weighted averaging.

\paragraph{Visible and hidden slots.}
Visible slots are initialized from fused observed object hypotheses. Hidden slots are initialized as persistent learned queries and allow the model to represent unobserved objects that may affect planning. We use $16$ hidden slots per scene. During training, hidden slots are matched to unobserved ground-truth objects when available; unmatched hidden slots are supervised toward low existence probability or zero displacement during world-model fine-tuning.

\paragraph{Slot transformer.}
Visible and hidden slots are refined using a slot transformer with model dimension $128$, $8$ attention heads, $4$ layers, feed-forward dimension $256$, and dropout $0.1$. Each layer performs self-attention over the combined slot set, followed by cross-attention to per-view shape and geometry tokens. The geometry-token dimension is $64$, the maximum number of observed tokens is $2048$, and learned positional embeddings are used for slot and view tokens.

\paragraph{Gaussian refinement and outputs.}
After slot refinement, each slot emits $8$ local Gaussians with anchor positions tied to the slot center. Gaussian scales are constrained to $[0.01,0.15]$ m, with maximum offset $0.08$ m and maximum anchor drift $0.16$ m. The Gaussian representation is rendered into a sparse occupancy volume at voxel size $0.015$ m over the shelf bounds with $0.05$ m padding. The final slot output is
\begin{equation}
    s_i^t = (\mu_i^t,\sigma_i^t,p_i^t),
\end{equation}
where $\mu_i^t\in\mathbb{R}^3$ is the center, $\sigma_i^t\in\mathbb{R}^3$ is the size, and $p_i^t\in[0,1]$ is the existence probability. Thus, each slot contains $7$ scalar values. The complete reconstructed state is $z_t=\{s_i^t\}_{i=1}^{K}$.

\paragraph{Relational graph.}
As shown in Fig.~\ref{fig:relational_graph}, GHOST produces a relational graph over object slots. Contact edges are added when object AABBs lie within $0.005$ m. Support edges require vertical contact and non-intersecting AABBs. Blocker edges are computed using projected overlap between an object and the target or goal corridor. We use a $5\%$ projected-overlap threshold for blocker edges.

\subsection{Relational Planning State and Metrics}
\label{app:planning_state_metrics}

\paragraph{Planning state.}
The relational planning state is derived as
\begin{equation}
    x_t = \mathcal{P}(z_t,g),
\end{equation}
where $z_t$ is the reconstructed slot state and $g$ is the task goal. Each object entry contains its id, category, center, size, visibility fraction, visible/hidden state, and target flag. The state also contains blocker, contact, and in-front-of relations, along with free-space estimates in the left, right, and back directions. The per-slot planning feature vector has length $13$ and includes normalized center, normalized size, visibility, target indicator, blocker indicator, directional free-space values, and a slot bias.

\paragraph{Blocker relation.}
A blocker relation from object $i$ to target $t$ is added if $i$ lies in front of $t$ along the shelf depth axis and overlaps the target or access corridor in the lateral-vertical cross-section. Specifically, object $i$ is considered in front of $t$ if
\begin{equation}
    x_i + 10^{-3} < x_t,
\end{equation}
and its projected YZ overlap exceeds $5\%$.

\paragraph{Visibility.}
The visibility score measures how much of the task site is visible from a fixed front-facing evaluation camera. For Reveal tasks, it is the visible fraction of the target object. During rollout evaluation, we approximate visibility by projecting objects in front of the target onto the target's YZ cross-section:
\begin{equation}
    \mathrm{Vis}(t) = 1 - \mathrm{clip}\left(\frac{A_{\mathrm{blocked}}(t)}{A_{\mathrm{YZ}}(t)},0,1\right).
\end{equation}
For Place tasks, visibility is computed over the goal region and normalized by the pixel area of the same region under an empty-shelf rendering.

\paragraph{Accessibility.}
Accessibility measures whether the task site can be reached from the shelf opening. For Reveal tasks, it is computed from side room, back room, and blocker count:
\begin{equation}
    \mathrm{Acc}(t) =
    \frac{
    0.5 \min(\mathrm{side\_room}/0.06,1)
    +
    0.5 \min(\mathrm{back\_room}/0.08,1)
    }{1+\mathrm{blocker\_count}(t)}.
\end{equation}
For Place tasks, accessibility is computed using a front-to-site corridor from the shelf opening to the goal-region center. The corridor has width $0.08$ m, height $0.12$ m, and depth equal to the distance from the shelf opening to the task site.

\paragraph{Success criteria.}
Reported success in Fig.~\ref{fig:results} is computed from simulator ground-truth geometry, while planning uses the reconstructed state. Reveal succeeds if
\begin{equation}
    \mathrm{Vis}(t) \geq 0.6,\quad
    \mathrm{Acc}(t) \geq 0.6,\quad
    \mathrm{blocker\_count}(t)=0.
\end{equation}
Reveal\_Multi requires this predicate for every target. Place succeeds if the target object center lies inside the assigned goal region $\mathcal{G}$:
\begin{equation}
    \mu_t \in \mathcal{G}.
\end{equation}
Place\_Multi requires all target objects to satisfy this predicate. Actions are marked invalid when the predicted transition violates shelf bounds, object existence, primitive validity, or geometric feasibility constraints.

\subsection{Action Space and LLM-Guided Action Proposals}
\label{app:action_llm}

\paragraph{Primitive actions.}
The action space contains four object-centric primitives:
\begin{equation}
    a=(i,u),\quad
    u\in\{\texttt{shift\_left},\texttt{shift\_right},\texttt{shift\_back},\texttt{pull\_forward}\},
\end{equation}
where $i$ is the object index. The primitive directions are defined in the shelf frame.

\begin{table}[ht]
    \centering
    \small
    \caption{Discrete manipulation primitives used by \method.}
    \label{tab:app_primitives}
    \begin{tabular}{l c c l}
        \toprule
        Primitive & Axis & Sign & Termination \\
        \midrule
        \texttt{shift\_left} & $y$ & $-1$ & wall contact \\
        \texttt{shift\_right} & $y$ & $+1$ & wall contact \\
        \texttt{shift\_back} & $x$ & $+1$ & wall contact \\
        \texttt{pull\_forward} & $x$ & $-1$ & front boundary \\
        \bottomrule
    \end{tabular}
\end{table}

The maximum primitive magnitude is $0.08$ m, the minimum motion threshold is $10^{-4}$ m, and the contact margin is $0.002$ m. The planner does not command object rotations; rotations may arise passively from physics.

\paragraph{LLM input and output.}
The LLM receives the task goal $g$ and the initial relational state $x_0$. The serialized input includes target id, target visibility/accessibility summary, object list, blocker relations, contact relations, in-front-of relations, free-space estimates, and valid primitive names. The LLM proposes compact object-direction candidates; it does not predict physical feasibility or choose the final plan. We use Qwen2.5-7B-Instruct as the open-source action-proposal model~\cite{yang2024qwen25}.

\begin{verbatim}
You are helping a robot plan in a cluttered shelf.
The robot can only use these object-centric primitives:
shift_left, shift_right, shift_back, pull_forward.

You are given a relational planning state with objects,
target information, blocker relations, visibility,
accessibility, and directional free-space estimates.

Your task is to propose compact action candidates for search.
Do not predict physics. Do not choose the final plan.
Only suggest which objects are worth considering and which
coarse direction is likely useful.

Return JSON only with this schema:
{
  "action_proposals": [
    {
      "object_ids": [int, ...],
      "direction": "shift_left|shift_right|shift_back|pull_forward",
      "reason": "short phrase"
    }
  ]
}

Rules:
- Prefer objects that block the target or the goal region.
- Prefer directions that open visibility or a front-facing access corridor.
- Include at most 3 action proposals.
- Each proposal may contain 1 to 4 object ids.
- Use only object ids that appear in the input.
- Use only the four allowed primitive names.

Input:
{TASK_AND_RELATIONAL_STATE_JSON}
\end{verbatim}

Each proposal is represented as $c=(\mathcal{I}_c,d_c)$, where $\mathcal{I}_c$ is a subset of object ids and $d_c$ is a coarse primitive direction. The expanded action set is
\begin{equation}
    \mathcal{A}(c)=\{(i,d_c)\mid i\in\mathcal{I}_c\}\cup\mathcal{A}_{\mathrm{aux}}.
\end{equation}
Auxiliary actions are generated deterministically from geometry, including pulling the target-facing blocker forward and shifting the nearest corridor blocker laterally toward the side with more free space. If parsing fails, the system falls back to these deterministic proposals. The LLM is served locally with temperature $0.2$, top-$p=0.9$, and a maximum of $512$ new tokens.

\subsection{World Model Training}
\label{app:world_model}

\paragraph{Inputs and targets.}
The learned world model receives object features, a padding mask, global action features, and a hallucinated-slot mask. Object features have dimension $13$ and include normalized center, normalized size, existence probability, visibility fraction, target flag, acted-object flag, hallucinated-slot flag, back-row flag, and a slot bias. Global features have dimension $11$ and include the acted-object summary, primitive one-hot encoding, and primitive direction vector.

The world model predicts per-object displacement, moved-object logits, and transition validity:
\begin{equation}
    f_{\phi}(z_t,a_t) \rightarrow \{(\Delta \mu_i, m_i, v)\}_{i=1}^{K},
\end{equation}
where $\Delta\mu_i\in\mathbb{R}^3$ is the normalized displacement, $m_i$ is a moved-object indicator, and $v$ is a transition-validity flag. Object sizes and existence probabilities are passed through unchanged.

\paragraph{Architecture.}
The world model uses an object-token encoder, a global-feature encoder, self-attention interaction layers, and prediction heads. The object encoder is a two-layer MLP with GELU activations and hidden dimension $96$. The global feature is encoded and broadcast to all object tokens. The interaction module contains $2$ transformer-style blocks with $4$ attention heads, LayerNorm, and feed-forward dimension $2\times96$. The trunk hidden dimension is $128$. The displacement head outputs a 3D displacement per slot, the moved head outputs one logit per slot, and the valid head outputs one transition-validity logit from the pooled global feature.

\paragraph{Loss.}
The world-model loss is
\begin{equation}
    \mathcal{L}_{\mathrm{WM}}
    =
    w_{\Delta}\mathcal{L}_{\Delta}
    +
    w_m\mathcal{L}_m
    +
    w_v\mathcal{L}_v
    +
    \mathbf{1}_{\{\mathrm{phase}=2\}} w_h\mathcal{L}_h.
\end{equation}
Here, $\mathcal{L}_{\Delta}$ is a masked Smooth-L1 displacement loss, $\mathcal{L}_m$ is binary cross-entropy for moved-object prediction, $\mathcal{L}_v$ is binary cross-entropy for transition validity, and $\mathcal{L}_h$ penalizes nonzero motion on hallucinated slots. We use weights $(w_{\Delta},w_m,w_v,w_h)=(1.0,0.5,0.25,0.5)$.

\paragraph{Training schedule.}
World-model training uses the transition dataset described above. Phase 1 trains on oracle simulator states, and Phase 2 fine-tunes on GHOST-reconstructed states while keeping oracle pre/post states for supervision. Phase 1 trains for $8$ epochs with batch size $32$ and learning rate $3\times10^{-4}$. Phase 2 initializes from Phase 1 and trains for $4$ epochs with batch size $32$ and learning rate $10^{-4}$. We use AdamW~\cite{loshchilov2019decoupled}, weight decay $10^{-5}$, gradient clipping at $5.0$, and fp32 loss accumulation.

\subsection{MCTS and Rollout Scoring}
\label{app:mcts}

\paragraph{Search tree.}
At each decision, the MCTS module~\cite{browne2012survey} in Fig.~\ref{fig:wm_generation} builds a new search tree from the current root state $z_0$. Each node stores the predicted world state, relational planning state, parent pointer, visit count, accumulated value, prior, depth, rollout summary, and applied action. Root actions are generated from the LLM-induced candidate set. Deeper nodes use deterministic geometric proposals and do not call the LLM again.

\paragraph{Selection and expansion.}
Selection uses a PUCT-style rule:
\begin{equation}
    a^* = \arg\max_a \left[
    Q(s,a) + c_{\mathrm{puct}}P(s,a)\frac{\sqrt{N(s)}}{1+N(s,a)}
    \right],
\end{equation}
with $c_{\mathrm{puct}}=1.25$. We use $32$ MCTS iterations, maximum depth $2$, branch factor $6$, rollout horizon $2$, and discount $0.97$. The full episode budget is $T=12$ actions.

\paragraph{Rollout score.}
Each rollout $\tau$ is summarized by visibility gain, accessibility gain, blocker change, and an invalid-transition flag:
\begin{equation}
    r(\tau) =
    \left(
    \Delta \mathrm{vis}(\tau),
    \Delta \mathrm{acc}(\tau),
    \Delta \mathrm{block}(\tau),
    I_{\mathrm{invalid}}(\tau)
    \right).
\end{equation}
The rollout score is
\begin{equation}
    V(\tau)
    =
    \Delta \mathrm{vis}
    +
    \Delta \mathrm{acc}
    -
    0.25\,B_{\mathrm{end}}
    -
    0.10\max(0,\Delta \mathrm{block})
    -
    1.0\,I_{\mathrm{invalid}},
\end{equation}
where $B_{\mathrm{end}}$ is the final blocker count. These terms correspond to the diagnostics reported in Table~\ref{tab:traj_metrics}. On one NVIDIA L40S GPU, the median LLM proposal latency is $0.42$ s, the mean MCTS wall time per planning step is $0.68$ s, and the mean world-model forward time is $4.8$ ms.

\subsection{Simulation Evaluation}
\label{app:simulation_eval}

Each simulation setting is evaluated over $500$ independent rollouts and $3$ random seeds. Results in Fig.~\ref{fig:results} are reported as means over seeds, with error bars denoting standard error. Random seeds affect scene sampling, object placement, camera subset selection, and stochastic LLM decoding. The default evaluation distribution uses $8$--$16$ objects, and multi-target tasks use two target objects.

\paragraph{Closed-loop sensing.}
The initial-view protocol is a controlled sparse-observation stress test.
For deployment, closed-loop sensing refreshes the scene reconstruction and
replans after each primitive. Table~\ref{tab:closed_loop} compares the two
protocols in simulation using the same primitive executor, $T=12$ budget,
and success criteria. Re-observing every two primitives gives
$84.8\pm3.1\%$ success and $4.6\%$ hidden-scene failures. Re-observing after
every primitive reduces hidden-scene failures from $7.6\%$ to $3.3\%$ of
all episodes, while episode time increases from $10.4$ to $24.2$ seconds.

\begin{table}[ht]
\centering
\small
\caption{Simulation success (\%) with initial observations only and
re-observation after every primitive.}
\label{tab:closed_loop}
\begin{tabular}{lcc}
\toprule
Method & Initial views only & Re-observation \\
\midrule
$\pi_{0.5}$ & 55.5 & $63.2\pm4.1$ \\
Long-VLA & 52.3 & $64.0\pm4.0$ \\
ParticleFormer & 72.8 &  $82.3\pm3.9$ \\
\method & $80.8\pm3.5$ & $86.3\pm3.0$ \\
\bottomrule
\end{tabular}
\end{table}

\paragraph{Proposal and search ablations.}
With the same world-model and MCTS budget, geometry-only proposals achieve
$74.6\pm3.9\%$ success, compared with $80.8\pm3.5\%$ for LLM plus geometry.
Local search depths $1$, $2$, $3$, and $4$ achieve $70.6\pm4.2\%$,
$80.8\pm3.5\%$, $81.5\pm3.4\%$, and $81.1\pm3.6\%$, respectively, with
relative world-model compute of $1.0\times$, $1.8\times$, $2.6\times$,
and $3.4\times$.

\paragraph{Predicted versus executed success.}
Across $500$ evaluation episodes, predicted-state and ground-truth success
agree in $89.2\%$ of cases; predicted-success precision is $91.7\%$,
and false positives account for $7.0\%$ of all episodes. Visibility and
accessibility MAE are $0.078$ and $0.094$, respectively. Replacing only
GHOST's initial state with oracle geometry raises success from
$80.8\pm3.5\%$ to $87.8\pm2.8\%$.

\paragraph{Reconstruction quality.}
On the held-out reconstruction set using Hungarian matching, multi-view
GHOST achieves $72.6\%$ hidden-object recall and $0.81$ blocker F1,
compared with $58.9\%$ and $0.71$ for single-view GHOST. These diagnostics
measure reconstruction separately from downstream task success.

\subsection{Real-World Hardware and Protocol}
\label{app:real_robot}
\begin{figure}[ht]
    \centering
    \includegraphics[width=0.9\textwidth]{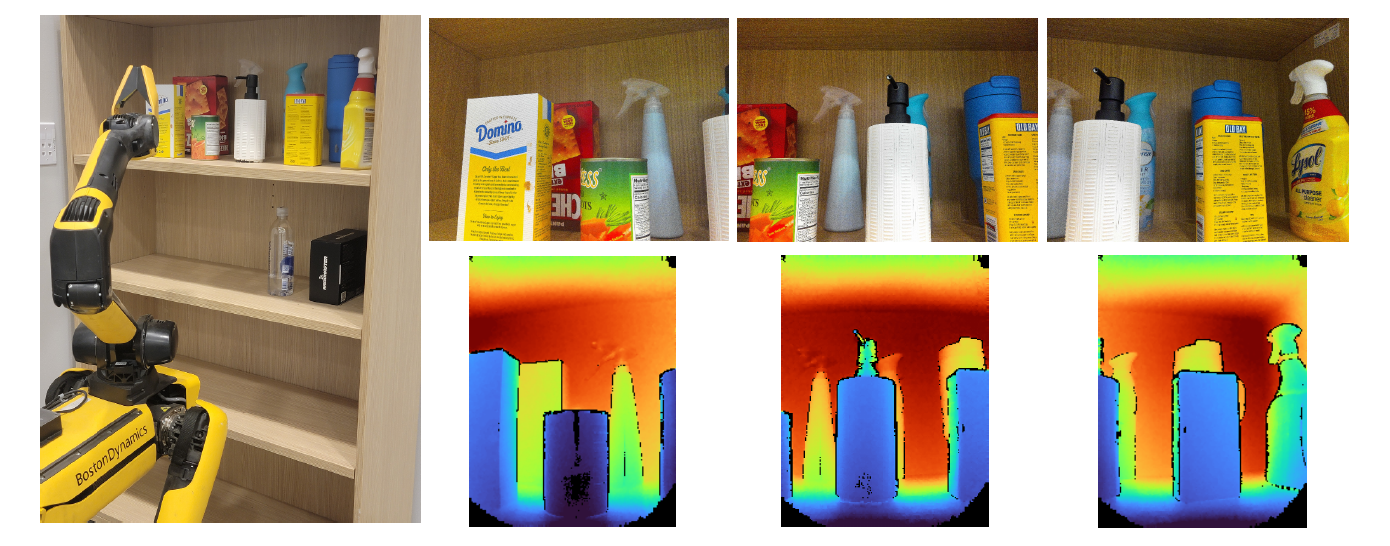}
    \caption{Real-world shelf observation setup. Left: Spot observes the cluttered shelf using its wrist-mounted camera. Right: representative robot-captured RGB views and corresponding depth maps used for scene reconstruction.}
\label{fig:app_real_observations}
\end{figure}
\paragraph{Robot platform.}
Real-world experiments in Fig.~\ref{fig:demo} are conducted on a Boston Dynamics Spot robot equipped with the Spot Arm~\cite{spot_arm_specs}. The Spot Arm is mounted in the standard configuration and is controlled through the Spot SDK. We use the standard Spot Arm gripper, which provides actuated gripper jaws and integrated sensing rather than a custom parallel-jaw end effector~\cite{spot_arm_specs}.

\paragraph{Observation.}
The real system uses the wrist-mounted gripper sensors. The Spot Arm gripper includes RGB and ToF depth sensing, which we use for frontal shelf observation~\cite{spot_arm_specs}. The RGB stream is captured at $3840\times2160$ resolution and resized for perception. The ToF depth stream is aligned to the RGB frame and filtered to the shelf workspace. We capture $3-5$ frontal wrist observation at the start of each planning episode. Fig.~\ref{fig:app_real_observations} shows the real shelf observation setup and representative robot-captured RGB-D views used for scene reconstruction.

\paragraph{Shelf setup.}
We construct real cluttered shelf environments using household objects. The real shelf has approximate dimensions $0.36$ m depth, $0.90$ m width, and $0.40$ m height, matching the simulated shelf scale. Each trial contains $6$--$10$ household objects, including bottles, cans, boxes, mugs, and containers.

\paragraph{Task setup.}
We evaluate Reveal and Place tasks. Reveal requires exposing and accessing a partially occluded target, while Place requires moving a specified object into a marked goal region after clearing nearby clutter. We run $10$ trials per task, for $20$ total trials. \method succeeds in $12/20$ trials, corresponding to $60\%$ overall success, with $7/10$ Reveal successes and $5/10$ Place successes.

\paragraph{Primitive execution and success.}
Each primitive is executed as a Cartesian end-effector motion in the shelf frame. Push primitives use a straight-line motion of $0.08$ m at approximately $0.05$ m/s. Pull primitives establish contact near the front face of the selected object and retract by $0.08$ m along the negative depth direction. After each primitive, the end effector retracts upward by $0.04$ m. The real-world action budget is $T=12$, matching simulation. Reveal succeeds if the target is exposed and reachable from the shelf opening; Place succeeds if the specified object is moved into the marked goal region. A trial is marked unsuccessful if the robot exhausts the action budget, executes an invalid primitive, or leaves the target or goal region blocked.

\paragraph{Real-world failures.}
Among the $8$ unsuccessful real-world trials, $3$ failures are due to execution errors such as slipping or insufficient object displacement, $3$ are due to perception or reconstruction errors, $2$ are due to transition-prediction mismatch.

\end{document}